\documentclass[11pt,a4paper]{article}
\usepackage[hyperref]{naaclhlt2019}
\usepackage{times}
\usepackage{latexsym}
\usepackage{graphicx}
\usepackage{subcaption}
\usepackage{amsmath}
\usepackage{amssymb}
\usepackage{booktabs}
\usepackage{multirow}
\usepackage{enumitem}

\usepackage{url}
\usepackage{xcolor}
\newcommand{\comment}[1]{}

\aclfinalcopy 

\newcommand{\bx}{\mathbf{x}}

\title{Cross-Lingual Alignment of Contextual Word Embeddings,  \\ with Applications to Zero-shot Dependency Parsing}

\comment{\author{First Author \\
  Affiliation / Address line 1 \\
  Affiliation / Address line 2 \\
  Affiliation / Address line 3 \\
  {\tt email@domain} \\\And
  Second Author \\
  Affiliation / Address line 1 \\
  Affiliation / Address line 2 \\
  Affiliation / Address line 3 \\
  {\tt email@domain} \\}
}

\author{Tal Schuster\thanks{\;\; Equal contribution}$^{*,1},$ ~~ Ori Ram$^{*,2},$ ~~
Regina Barzilay$^1,$ ~~
Amir Globerson$^2$ \\
\mbox{}\\
$^1$Computer Science and Artificial Intelligence Lab, MIT \\
$^2$Tel Aviv University \\
\small{\texttt{\{tals, regina\}@csail.mit.edu},~~~~\texttt{\{ori.ram, gamir\}@cs.tau.ac.il}}}

\date{}

\newcommand{\vv}{\boldsymbol{v}}
\newcommand{\ee}{\boldsymbol{e}}

\newcommand{\bb}{\boldsymbol{b}}
\newcommand{\hh}{\boldsymbol{h}}
\newcommand{\uu}{\boldsymbol{u}}

\newcommand{\dict}{D}
\newcommand{\secref}[1]{Section \ref{#1}}
\newcommand{\tabref}[1]{Table \ref{#1}}
\newcommand{\figref}[1]{Figure \ref{#1}}
\renewcommand{\eqref}[1]{Eq.~\ref{#1}}
\newcommand{\appref}[1]{App.~\ref{#1}}

\begin{document}
\maketitle

\begin{abstract}
We introduce a novel method for multilingual transfer that utilizes deep contextual embeddings, pretrained in an unsupervised fashion.
While contextual embeddings have been shown to yield richer representations of meaning compared to their static counterparts, aligning them poses a challenge due to their dynamic nature.
To this end, we construct context-independent variants of the original monolingual spaces and utilize their mapping to derive an alignment for the context-dependent spaces. 
This mapping readily supports processing of a target language, improving transfer by context-aware embeddings.
Our experimental results demonstrate the effectiveness of this approach for zero-shot and few-shot learning of dependency parsing. Specifically, our method consistently outperforms the previous state-of-the-art on 6 tested languages, yielding an improvement of 6.8 LAS points on average.\footnote{Code and models: \url{https://github.com/TalSchuster/CrossLingualELMo}.}
\end{abstract}
\section{Introduction}

Multilingual embedding spaces have been demonstrated to be a promising means for enabling cross-lingual transfer in many natural language processing tasks (e.g.\ \citet{ammar_one_2016, lample2018unsupervised}). Similar to how universal part-of-speech tags enabled parsing transfer across languages~\cite{petrov2012universal}, multilingual word embeddings further improve transfer capacity by enriching models with lexical information.
Since this lexical representation is learned in an unsupervised fashion and thus can leverage large amounts of raw data, it can capture a more nuanced representation of meaning than unlexicalized transfer. Naturally, this enrichment is translated into improved transfer accuracy, especially in low-resource scenarios \cite{guo2015cross}. 

In this paper, we are moving further along this line and exploring the use of contextual word embeddings for multilingual transfer. 
By dynamically linking words to their various contexts, these embeddings provide a richer semantic and syntactic representation than traditional context-independent word embeddings \citep{peters_deep_2018}. A straightforward way to utilize this richer representation is to directly apply existing transfer algorithms on the contextual embeddings instead of their static counterparts. 
In this case, however, each token pair is represented by many different vectors corresponding to its specific context. 
Even when supervision is available in the form of a dictionary, it is still unclear how to utilize this information for multiple contextual embeddings that correspond to a word translation pair.

In this paper, we propose a simple but effective mechanism for constructing a multilingual space of contextual embeddings. 
Instead of learning the alignment in the original, complex contextual space, we drive the mapping process using context-independent embedding anchors. We obtain these anchors by factorizing the contextual embedding space into context-independent and context-dependent parts. 
Operating at the anchor level not only compresses the space, but also enables us to utilize a word-level bilingual dictionary as a source of supervision, if available. Once the anchor-level alignment is learned, it can be readily applied to map the original spaces with contextual embeddings. 

Clearly, the value of word embeddings depends on their quality, which is determined by the amount of raw data available for their training \cite{jiang2018learning}. We are interested in expanding the above approach to the truly low-resource scenario, where a language  not only lacks annotations, but also has limited amounts of raw data. In this case, we can also rely on a data rich language to stabilize monolingual embeddings of the resource-limited language. As above, context-independent anchors are informing this process.  Specifically, we introduce an alignment component to the loss function of the language model, pushing the anchors to be closer in the joint space. While this augmentation is performed on the static anchors, the benefit extends to the contextual embeddings space in which we operate.

We evaluate our aligned contextual embeddings on the task of zero-shot cross-lingual dependency parsing. Our model consistently outperforms previous transfer methods, yielding absolute improvement of 6.8 LAS points over the prior state-of-the-art~\cite{ammar_one_2016}. We also perform comprehensive studies of simplified variants of our model.
Even without POS tag labeling or a dictionary, our model performs on par with context-independent models that do use such information. Our results also demonstrate the benefits of this approach for few-shot learning, i.e.\ processing languages with limited data. Specifically, on the Kazakh tree-bank from the recent CoNLL 2018 shared task with only 38 trees for training, the model yields 5 LAS points gain over the top result~\cite{smith201882}.

\section{Related work}

\paragraph{Multilingual Embeddings} The topic of cross-lingual embedding alignment is an active area of research~\cite{mikolov2013exploiting, xing2015normalized, dinu2014improving, lazaridou2015hubness, zhang2017adversarial}. Our work most closely relates to MUSE~\cite{conneau2017word}, which constructs a multilingual space by aligning monolingual embedding spaces. When a bilingual dictionary is provided, their approach is similar to those of \cite{smith2017offline, artetxe2017learning}. MUSE extends these methods to the unsupervised case by constructing a synthetic dictionary.
The resulting alignment achieves strong performance in a range of NLP tasks, from sequence labeling~\cite{lin18multi} to natural language inference~\cite{Conneau18xnli} and machine translation~\cite{lample2018unsupervised, Qi18when}. 
 Recent work further improves the performance on both the supervised \cite{joulin2018loss} and unsupervised \cite{grave2018unsupervised, alvarez2018gromov, hoshen2018non} settings for context-independent embeddings.

While MUSE operates over token based embeddings, we are interested in aligning contextual embeddings, which have shown their benefits in several monolingual applications~\cite{peters_deep_2018, mccann_cove_2017, howard_ulmfit_2018, radford_improving_2018, devlin_bert_2018}. 
However, this expansion introduces new challenges which we address in this paper. 

In a concurrent study, \citet{aldarmaki2019context} introduced an alignment that is based only on word pairs in the same context, using parallel sentences. Our method achieves better word translations without relying on such supervision.


Our work also relates to prior approaches that utilize bilingual dictionaries to improve embeddings that were trained on small datasets. For instance, \citet{xiao2014distributed} represent word pairs as a mutual vector, while \citet{adams2017cross} jointly train cross-lingual word embeddings by replacing the predicted word with its translation.  To utilize a dictionary in the contextualized case, we include a soft constraint that pushes those translations to be similar in their context-independent representation. A similar style of regularization was shown to be effective for cross-domain transfer of word embeddings~\citep{yang2017simple}.

\paragraph{Multilingual Parsing}
In early work on multilingual parsing, transfer was commonly implemented using delexicalized representation such as part-of-speech tags~\cite{mcdonald2011multi, petrov2012universal, naseem2012selective, tiedemann2015cross}.  

Another approach for cross-lingual parsing includes annotation projection and treebank translation~\cite{xiao2015annotation,wang_galactic_2016,Tiedemann2017cross}, which mostly require some source of supervision.

Advancements in multilingual word representations opened a possibility of lexicalized transfer. Some of these approaches start by aligning monolingual embedding spaces~\cite{zhang_hierarchical_2015, guo2015cross, guo_representation_2016, ammar_one_2016}, and using resulting word embeddings as word representations instead of universal tags. Other approaches are learning customized multilingual syntactic embeddings bootstrapping from universal POS tags~\cite{duong2015cross}. While some models also learn a language embedding~\citep{ammar_one_2016, de_lhoneux2018parameter}, it is unfeasible in a zero-shot scenario.

In all of the above cases, token-level embeddings are used. Inspired by strong results of using contextualized embeddings in monolingual parsing~\cite{che_towards_2018, wang2018improved, clark2018semi}, we aim to utilize them in  the multilingual transfer case. 
Our results demonstrate that richer representation of lexical space does lead to significant performance gains.

\section{Aligning Contextual Word Embeddings}\label{section:alignment}

\begin{table}[t]
\centering
\begin{tabular}{ccc}
\toprule
\multirow{2}{*}{$D_{cos}(\bar{\ee}_i, \bar{\ee}_j) $} & \multicolumn{2}{c}{$D_{cos}(\bar{\ee}_i, \ee_{i,c}) $}  \\
& \textsc{All Words} & \textsc{Homonyms} \\
\midrule
0.85 ($\pm 0.09$)	& 0.18 ($\pm 0.04$)	& 0.21 ($\pm 0.04$)  \\ 
\bottomrule
\end{tabular}
\caption{Average cosine distances between pairs of embedding anchors (left column) and between contextualized embeddings of words to their corresponding anchor. The right column includes these distances only for homonyms, whereas the center column is averaged across all words. Only alphabetic words with at least 100 occurrences were included.}\label{tab:dists}
\end{table}

In this section we describe several approaches for aligning context-dependent embeddings from a source language $s$ to a target language $t$. We address multiple scenarios, where different amounts of supervision and data are present.
Our approach is motivated by interesting properties of context-dependent embeddings, which we discuss later.

We begin with some notations:
\begin{itemize}
\item {\bf Context Dependent Embeddings:} Given a context $c$ and a token $i$, we denote the embedding of $i$ in the context $c$ by $\ee_{i,c}$. We use $\ee_{i,\cdot}$ to denote the point cloud of all contextual embeddings for token $i$.
\item {\bf Embedding Anchor:}
Given a token $i$ we denote the anchor of its context dependent embeddings by $\bar{\ee}_i$, where:
\begin{equation}
\bar{\ee}_i = \mathbb{E}_c\big[\ee_{i,c}\big]\ .
\end{equation}
In practice, we calculate the average over a subset of the available unlabeled data.
\item {\bf Shift From Mean:}
For any embedding $\ee_{i,c}$ we can therefore define the shift $\hat{\ee}_{i,c}$ from the average via:
\begin{equation}
\label{eq:emb_break}
    \ee_{i,c} = \bar{\ee}_i + \hat{\ee}_{i,c}\ \ .
\end{equation} 

\item {\bf Embedding Alignment:} Given an embedding $\ee^s_{i,c}$ in $s$, we want to generate an embedding $\ee^{s\to t}_{i,c}$ in the target language space, using a linear mapping $W^{s\rightarrow t}$. Formally, our alignment is always of the following form:
\begin{equation}\label{eq:alignment}
\ee^{s\to t}_{i,c} = W^{s\rightarrow t} \ee^s_{i,c}\ \ .
\end{equation}


\end{itemize}

\subsection{The Geometry of Context-Dependent Embeddings \label{sec:geometry}}
A given token $i$ can generate multiple vectors $\ee_{i,c}$, each corresponding to a different context $c$.
A key question is how the point cloud  $\ee_{i,\cdot}$ is distributed. In what follows we explore this structure, and reach several conclusions that will motivate our alignment approach. The following experiments are performed on ELMo~\cite{peters_deep_2018}.

\begin{figure}[t!]
    \centering
  \includegraphics[width=0.49\textwidth]{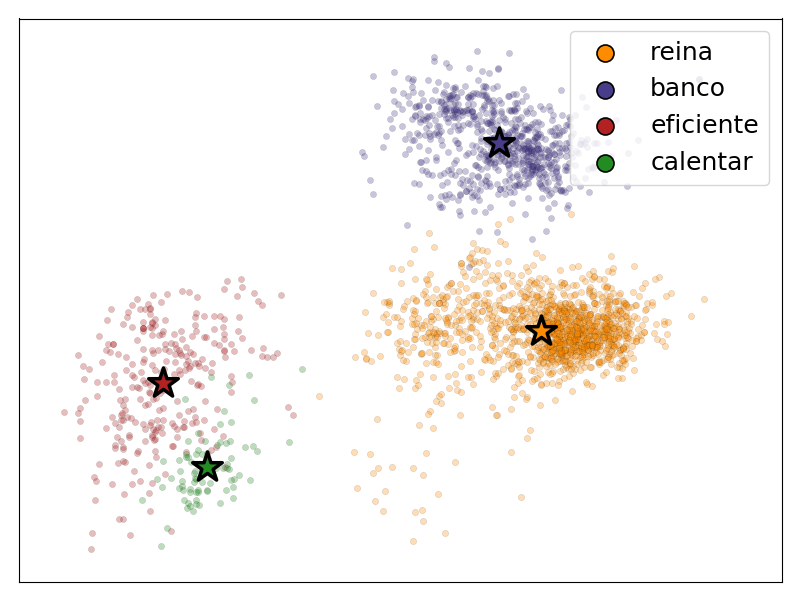}

    \caption{A two dimensional PCA showing examples of contextual representations for four Spanish words. Their corresponding anchors are presented as a star in the same color. (best viewed in color)} \label{fig:emb_spaces}
\end{figure}

\comment{
\begin{table}[t]
\centering
\begin{tabular}{l|cc}
\toprule
\textsc{Language}    & $ D_{cos}(\bar{\ee}_i, \ee_{i,c}) $   & $D_{cos}(\bar{\ee}_i, \bar{\ee}_j) $   \\ 
\midrule
\textsc{English}	& 0.18 (0.04)	& 0.85 (0.09) \\ 
\textsc{German}	& 0.37 (0.14)	& 0.88 (0.09) \\ 
\textsc{Italian}	& 0.33 (0.16)	& 0.87 (0.08) \\ 
\textsc{Swedish}	& 0.32 (0.23)	& 0.91 (0.09) \\ 

\bottomrule
\end{tabular}
\caption{Average (std) cosine distances between the contextualized embeddings of words to the mean vector per token (middle column), and between different mean vectors (right column). Only alphabetic words with at least 100 occurrences were included.}\label{tab:dists}
\end{table} 
}

{\bf Point Clouds are Well Separated} A cloud   $\ee_{i,\cdot}$ corresponds to occurrences of the word $i$ in different contexts. Intuitively, we would expect its points to be closer to each other than to points from  $\ee_{j,\cdot}$ for a different word $j$. Indeed, when measuring similarity between points $\ee_{i,c}$ and their anchor $\bar{\ee}_i$, we find that these are much more similar than anchors of different words  $\bar{\ee}_i$ and $\bar{\ee}_j$ (see \tabref{tab:dists}). This observation supports our hypothesis that anchor-driven alignment can guide the construction of the alignment for the contextual space. 
A visualized example of the contextualized representations of four words is given in Figure~\ref{fig:emb_spaces}, demonstrating the appropriateness of their anchors.
Still, as previous studies have shown, and as our results point, the context component is very useful for downstream tasks.



\comment{
\begin{table}[t]
\centering
\begin{tabular}{ll|ll}
\toprule
\multicolumn{2}{c|}{\textsc{Bottom 10}} & \multicolumn{2}{c}{\textsc{Top 10}} \\
\midrule
disambiguation & 0.05    & as  & 0.39 \\
householder    & 0.06     & on  & 0.4  \\
depending      & 0.06      & a   & 0.4  \\
decides        & 0.07  & for & 0.41 \\
redirect       & 0.07  & or  & 0.41 \\
refer          & 0.07    & in  & 0.43 \\
asks           & 0.08   & the & 0.46 \\
discovers      & 0.08 & to  & 0.46 \\
unable         & 0.08  & of  & 0.5  \\
sentenced      & 0.08 & and & 0.51 \\
\bottomrule
\end{tabular}
\caption{ Examples of English words sorted by the average cosine distance of the contextual representations to their mean $ D_{cos}(\bar{\ee}_i, \ee_{i,c}) $. Only words that had at least two thousand occurrences in our English dev set were included.\protect\footnotemark
} \label{tab:word_dists}
\end{table}
\footnotetext{Numbers and punctuation had average large cosine distances but were excluded from the list.}
}

\begin{figure}[t]
\centering

  \includegraphics[width=0.49\textwidth]{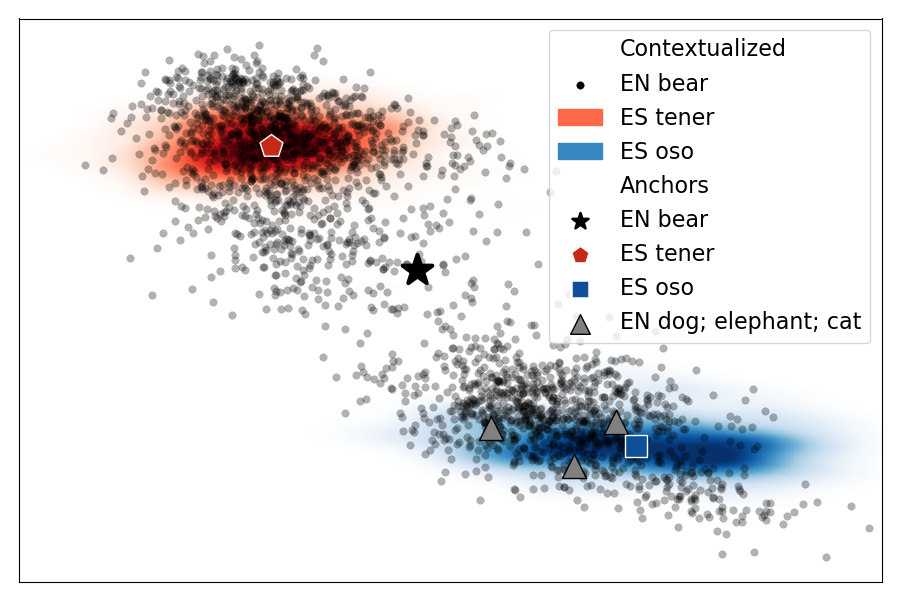}
  \caption{Contextual embeddings for the English word ``bear'' and its two possible translations in Spanish ---  ``oso'' (animal) in blue and  ``tener'' (to have) in red. The figure shows a two dimensional PCA for the aligned space of the two languages. The symbols are the anchors, the clouds represent the distribution of the contextualized Spanish words, and the black dots are for contextualized embeddings of ``bear''. The gray colored triangles show the anchors of the English words ``dog'', ``elephant'', ``cat'', from left to right respectively. 
}   \label{fig:bear}
\end{figure}

{\bf Homonym Point Clouds are Multi-Modal} When a word $i$ has multiple distinct senses, we might expect the embeddings for $i$ to reflect this by separating into multiple distinct clouds, one for each
meaning. Figure \ref{fig:bear} demonstrates that this indeed happens for the English word ``bear''. Furthermore, it can be seen that after alignment (\secref{sec:supervised}) with Spanish, the distinct point clouds are aligned with their corresponding distinct words in Spanish. See \appref{appendix:space} for another example.

We examined the shift from mean for a list of 250 English homonyms from Wikipedia.\footnote{\url{https://en.wikipedia.org/wiki/List_of_true_homonyms}} As \tabref{tab:dists} shows, the shift of these words is indeed slightly higher than it is for other words. However, they still remain relatively close to their per-token anchor. Therefore, these anchors can still serve as a good approximation for learning alignments.

\subsection{Context-Independent Alignment \label{sec:mikolov}}

We begin by briefly reviewing previous approaches for 
aligning context-independent embeddings, as they are generalized in this work to the contextual case. We denote the embedding of a word $i$ by $\ee_i$. At first, assume we are given word pairs $\{(\ee^s_i,\ee^t_i)\}$ from a source language $s$ and a target language $t$, and we look for a mapping between those.  \citet{mikolov2013exploiting} proposed to learn a linear transformation whereby $\ee^t_i$ is approximated via $W\ee^s_i$, for a learned matrix $W$. We focus on methods that follow this linear alignment.
The alignment matrix is found by solving:
\begin{equation}\label{eq:linear_mapping_independent}
    W^{s\rightarrow t} = \underset { W \in O _ { d } ( \mathbb { R } ) } { \operatorname { argmin } } \sum_{i=1}^{n} \Big\Vert  W \ee^s_i - \ee^t_i \Big\Vert  ^2\ ,
\end{equation}

\noindent where $O _ { d } ( \mathbb { R } )$ is the space of orthogonal matrices. This constraint was proposed by \citet{xing2015normalized}  in order to preserve inter-lingual relations.
Under this constraint, \eqref{eq:linear_mapping_independent} is an instance of the orthogonal Procrustes problem, which has a closed-form solution $W^{s\rightarrow t} = U V^{T}$. The columns of $U$ and $V$ are the left and right singular vectors of the multiplication of the source and (transposed) target embedding matrices.


For the \emph{unsupervised} case (i.e.\ when a dictionary is absent), \citet{conneau2017word} (MUSE) suggested to learn the alignment via adversarial training, such that a discriminator is trained to distinguish between target and aligned source embeddings. Thereafter, a \emph{refinement} procedure is applied iteratively as follows. First, a dictionary is built dynamically using the current alignment such that only words with high confidence are considered. Using the dictionary, the alignment matrix is re-calculated as in the supervised case.

\subsection{Context-Dependent Alignment} \label{sec:supervised}

We next turn our attention to the main task of this paper, which is aligning context-dependent embeddings. We now describe our generalization of the methods described in \secref{sec:mikolov} for this case. The first two methods are based only on anchors while the third one uses the contextual vectors themselves. Altogether, we suggest three alignment procedures, one aimed for the supervised and two for the unsupervised cases.

\paragraph{Supervised Anchored Alignment} As a first step, we are assuming access to a dictionary for the source and target domains. For each source word $i$ denote by $\dict(i)$ the corresponding word in the target language.\footnote{In practice, we may have multiple target words for a single source word, and the extension is straight-forward.}

In the context-dependent case, \eqref{eq:linear_mapping_independent} is no longer well-defined, as there are many corresponding vectors to both the source and the target words.
However, this challenge can be  addressed by aligning the vectors $\bar{\ee}_i$ for which we do have one per word. This is motivated by our observations in~\secref{sec:geometry} that context-dependent embeddings are well clustered around their centers.

Thus, in the case where a dictionary is available, we solve \eqref{eq:linear_mapping_independent} with token anchors as inputs.
\comment{

we shall obtain the alignment as follows. First, we solve the following optimization problem:

\begin{equation}
 W^{s\rightarrow t} =       \underset { W \in O _ { d } ( \mathbb { R } ) } { \operatorname { argmin } } \sum_{i=1}^{n} \Big\Vert  W \bar{\ee}^s_i - \bar{\ee}^t_{\dict(i)} \Big\Vert  ^2
\end{equation}
Thereafter, we use the above matrix to obtain target-language embeddings via \eqref{eq:alignment}.

}

We emphasize that by constraining $W^{s\rightarrow t}$ to be orthogonal, we also preserve relations between $\hat{\ee}_{i,c}$ and $\hat{\ee}_{i,c'}$ that represent the contextual information.



\paragraph{Unsupervised Anchored Alignment}
In this setting, no dictionary is present.
As in the supervised case, we can naturally extend a context-independent alignment procedure to the contextual space by leveraging the anchor space $\bar{\ee}_i$. This can be done using the adversarial MUSE framework proposed by \citet{conneau2017word} and described at the end of \secref{sec:mikolov}.

\paragraph{Unsupervised Context-based Alignment}
Alternatively, the alignment could be learned directly on the contextual space. To this end, we follow again the adversarial algorithm of MUSE, but for each word we use multiple embeddings induced by different contexts, rather than the word anchor.

This context-based alignment presents opportunities but also introduces certain challenges. 
On the one hand, it allows to directly handle homonyms during the training process. However, empirically we found that training in this setting is less stable than unsupervised anchored alignments.

\paragraph{Refinement}
As a final step, for both of the unsupervised methods, we perform the refinement procedure that is incorporated in MUSE (end of \secref{sec:mikolov}). In order to synthesize a dictionary, we use distance in the anchor space.

\subsection{Learning Anchored Language Models} \label{sec:anchored_lm}
Thus far we assumed that embeddings for both source and target languages are pretrained separately. Afterwards, the source is mapped to the target in a second step via a learned mapping. However, this approach may not work well when raw data for the source languages is scarce, resulting in deficient embeddings. In what follows, we show how to address this problem when a dictionary is available. We focus on embeddings that are learned using a language model objective but this can be easily generalized to other objectives as well.

Our key idea is to constrain the embeddings across languages such that word translations will be close to each other in the embedding space.  This can serve as a regularizer for the resource-limited language model. In this case, the anchors are the model representations prior to its context-aware components (e.g., the inputs to ELMo's LSTM). 

Denote the anchor for word $i$ in language $s$ by $\vv_i^s$. Now, assume we have trained a model for the target language and similarly have embeddings  $\vv_i^t$. We propose to train the source model with an added regularization term as follows:
\begin{equation}
    \lambda_{\text{anchor}}\cdot\sum_i \|\vv_i^s - \vv_{D(i)}^t\|_2^2\ ,
\end{equation}
where $\lambda_{\text{anchor}}$ is a hyperparamter. 
This regularization has two positive effects. First, 
it reduces overfitting by reducing the effective number of parameters the model fits (e.g., if the regularizer has large coefficient, these parameters are essentially fixed). Second, it provides a certain level of alignment between the source and target language since they are encouraged to use similar anchors.


\section{Multilingual Dependency Parsing}\label{section:dependency_parsing}

Now that we presented our method for aligning contextual embeddings, we turn to evaluate it on the task of cross-lingual dependency parsing. We first describe our baseline model, and then show how our alignment can easily be incorporated into this architecture to obtain a multilingual parser.

 \paragraph{Baseline Parser} Most previous cross-lingual dependency parsing models used transition-based models~\cite{ammar_one_2016, guo_representation_2016}. We follow \citet{che_towards_2018, wang2018improved, clark2018semi} and use a first-order graph-based model. Specifically, we adopt the neural edge-scoring architecture from \citet{dozat_deep_2017, dozat_stanford_2018}, which is based on \citet{kiperwasser_simple_2016}.
\comment{
Formally, let $s$ be a sentence with $n$ words, $\ee_{i}$ be the word embedding, and $\mathbf{p}_i$ be the POS-tag\footnote{We use only Universal POS tags.} embedding of the $i$-th word in $s$. The representation vector for each token is computed by:
\begin{equation}\label{eq:parsing_embeddings}
\begin{split}
\bx_i &= \text{concat} (\ee_{i}\ ,\ \mathbf{p}_i),\\
\mathbf{r}_i &= \text{Bi-LSTM}(\{\bx_1,...,\bx_n\})_i\ ,\\
\end{split}
\end{equation}
}
We now briefly review this architecture. Given a sentence $s$, let $\ee_i$ and $\mathbf{p}_i$ be its word and POS-tag embeddings. These are concatenated and fed into a Bi-LSTM to produce token-level contextual representations $\mathbf{r}_i$.
 Four Multi-Layer Perceptrons are applied on these vectors, resulting in new representations $\hh_i^{arc-dep}$, $\hh_i^{arc-head}$, $\hh_i^{rel-dep}$ and $\hh_i^{rel-head}$ for each word $i$. Arc scores are then obtained by:
\begin{equation}
s_{ij}^{arc} = \left(\hh_i^{arc-head}\right)^T \left(U^{arc} \hh_j^{arc-dep} + \bb^{arc}\right)\ .
\end{equation}
Additionally, the score for predicting the dependency label $r$ for an edge $(i,j)$ is defined as 
\begin{equation}
\begin{split}
s_{(i,j),r}^{rel} = &\left( \hh_i^{rel-head}\right)^T U_r^{rel} \hh_j^{rel-dep} \ + \\
&\left( \uu_r^{rel-head}\right)^T\hh_i^{rel-head} \ + \\
&\left( \uu_r^{rel-dep}\right)^T\hh_j^{rel-dep} + b_r \ .
\end{split}
\end{equation}

At test time, MST is calculated to ensure valid outputs. 
\paragraph{Multilingual Parsing with Alignment}
We now extend this model, in order to effectively use it for transfer learning. First, we include contextualized word embeddings by replacing the static embeddings with a pre-trained ELMo \cite{peters_deep_2018} model (instead of $\ee_i$). Second, we share all model parameters across languages and use the contextual word embeddings after they are aligned to a joint space $J$. 
Formally, if $s$ is a sentence of language $\ell$, contextual word embeddings are obtained via:
\begin{equation}
\ee_{i,s}^{\ell\rightarrow J} = W^{\ell \to J} \ee_{i,s}\ \ ,
\end{equation}
where $W^{\ell \to J}$ is the alignment matrix from language $\ell$ to the joint space.\footnote{We use the space of the training language as our joint space and align the tested language to it. In the multi-source scenario, we align all embeddings to English.} This alignment is learned apriori and kept fixed during parser training. This setup is applicable for both single and multiple training languages. For the tested language, training data could be available, sometimes limited (few-shot), or absent (zero-shot). The alignment methods are described in detail in \secref{section:alignment}.

In their paper, \citet{peters_deep_2018} suggest to output a linear combination over the representations of each layer of ELMo, learning these weights jointly with a downstream task. Our alignment is learned separately for each layer. Therefore, we keep the weights of the combination fixed during the training to ensure that the parser's inputs are from the joint cross-lingual space. 
Alternatively, one can share the weights of the combination between the languages and learn them.

All the above modifications are at the word embedding level, making them applicable to any other NLP model that uses word embeddings. 


\section{Experimental Setup}

\begin{table*}[t!]
\centering
\begin{tabular}{l|cccccc|c}
\toprule
\textsc{Alignment Method}       & \textsc{de}    & \textsc{es}    & \textsc{fr}    & \textsc{it}    & \textsc{pt}    & \textsc{sv}    & \textsc{average} \\ 
\midrule
\textsc{Supervised Anchored}                     & \ \ 78\ \  & \ \ 85\ \  & \ \ 86\ \  & \ \ 82\ \  & \ \ 74\ \  & \ \ 68\ \  &  \ \ 79\ \    \\ \midrule

\textsc{Unsupervised Anchored}  & 63 & 61 & 70 & 58 & 35 & 22 &   52  \\ 
\hspace{22mm} \textsc{ + Refine}                 & 72 & 74 & \textbf{81} & \textbf{77} & 53 & \textbf{33} &  \textbf{65}   \\ 

\textsc{Unsupervised Context-based} & 57 & 68 & 59 & 57 & 53 & * &   49  \\ 
\hspace{22mm} \textsc{ + Refine} & \textbf{73} & \textbf{82} & 77     & 73 & \textbf{66}     & *   & 62    \\ 
\bottomrule
\end{tabular}\\
\caption{Word translation to English precision @5 using CSLS~\cite{conneau2017word} with a dictionary (supervised) and without (unsupervised) for German (\textsc{de}), Spanish (\textsc{es}), French (\textsc{fr}), Italian (\textsc{it}), Portuguese (\textsc{pt}) and Swedish (\textsc{sv}). Each of the unsupervised results is followed by a line with the results post the anchor-based refinement steps. $\quad$ * stands for 'Failed to converge'. }\label{tab:align}
\end{table*}

\paragraph{Contextual Embeddings}
We use the ELMo model \cite{peters_deep_2018} with its default parameters to generate embeddings of dimension 1024 for all languages. For each language, training data comprises Wikipedia dumps\footnote{\url{https://lindat.mff.cuni.cz/repository/xmlui/handle/11234/1-1989}} that were tokenized using UDpipe~\cite{udpipe_2017}. We randomly shuffle the sentences and, following the setting of ELMO, use 95\% of them for training and 5\% for evaluation.

\paragraph{Alignment}
We utilize the MUSE framework\footnote{\url{https://github.com/facebookresearch/MUSE/}} \cite{conneau2017word} and the dictionary tables provided by them. 
The $\bar{\ee}_i$ (anchor) vectors for the alignment are generated by computing the average of representations on the evaluation set (except for the limited unlabeled data case). To evaluate our alignment, we use the anchors to produce word translations. For all experiments we use the 50k most common words in each language.

\begin{table*}[t!]
\begin{tabular}{l|cccccc|c}
\toprule
\textsc{Model}              & \textsc{de}        & \textsc{es}        & \textsc{fr}        & \textsc{it}        & \textsc{pt}        & \textsc{sv}        & \textsc{average}       \\ 
\midrule
\citet{zhang_hierarchical_2015}           & 54.1     & \ \ 68.3\ \      & 68.8     & 69.4     & 72.5     & 62.5     & 65.9     \\ 
\citet{guo_representation_2016}                & 55.9     & 73.1     & 71.0     & 71.2     & 78.6     & 69.5     & 69.9     \\ 
\citet{ammar_one_2016}           & \ \ 57.1\ \  & \ \ 74.6\ \  & \ \ 73.9\ \  & \ \ 72.5\ \  & \ \ 77.0\ \ & \ \ 68.1\ \  & 70.5 \\ 
\midrule 
\textsc{Aligned fastText} & 61.5 & 78.2  & 76.9 & 76.5 & \textbf{83.0} & 70.1 & 74.4 \\ 
\textsc{Aligned} $\bar{\ee}$       & 58.0 & 76.7 & 76.7 & 76.1 & 79.2 & 71.9 & 73.1 \\ 
\textsc{Ours}               & \textbf{65.2} & \textbf{80.0} & \textbf{80.8} & \textbf{79.8} & 82.7 & \textbf{75.4} & \textbf{77.3} \\ 
\midrule \midrule
\textsc{Ours, no dictionary} &    64.1 & 77.8 & 79.8 & 79.7 & 79.1 & 69.6 & 75.0 \\ 
\textsc{Ours, no pos}     & 61.4 & 77.5  & 77.0 & 77.6 & 73.9 &   71.0 & 73.1  \\ 
\textsc{Ours, no dictionary, no pos} & 61.7  &	76.6  &	76.3  &	77.1  &	69.1  &	54.2 & 69.2\\

\bottomrule
\end{tabular}
\caption{Zero-shot cross lingual LAS scores compared to previous methods, for German (\textsc{de}), Spanish (\textsc{es}), French (\textsc{fr}), Italian (\textsc{it}), Portuguese (\textsc{pt}) and Swedish (\textsc{sv}). Aligned \textsc{fastText} and $\bar{\ee}$ context-independent embeddings are also presented as baselines. The bottom three rows are models that don't use POS tags at all and/or use an unsupervised anchored alignment. Corresponding UAS results are provided in \appref{appendix:parsing}.}\label{table:zeroshot}
\end{table*}

\paragraph{Dependency Parsing}
We used the biaffine parser implemented in AllenNLP \cite{gardner_allennlp_2018}, refactored to handle our modifications as described in \secref{section:dependency_parsing}.\footnote{\url{https://github.com/TalSchuster/allennlp-MultiLang}} The parser is trained on trees from a single or multiple languages, as described in each setting (\secref{sec:results}). For the multiple case, we randomly alternate between the available languages, i.e.\ at each iteration we randomly choose one language and sample a corresponding batch.
Dropout \cite{srivastava_dropout:_2014} is applied on ELMo representations, Bi-LSTM representations and outputs of MLP layers. We also apply early stopping, where validation accuracy is measured as average LAS score on the development set across all training languages. 
The parser hyperparameters are the same as \citet{dozat_stanford_2018} except we reduce the POS tag embedding size from 100 to 50 and increase the head/dependent MLP dimension from 400 to 500. 
All hyperparameter values used are listed in \appref{appendix:hyperparameters}.  

From experiments on the English tree-bank, we found that using the outputs of the first LSTM layer is as good as learning a combination.\footnote{This was concurrently justified by \citet{liu-naacl}, showing that the first layer alone can perform better than a mixture.} This agrees with \citet{belinkov2017evaluating}, showing that lower layers capture more syntactic information. Therefore, we fix the weights over ELMo layers to $[0,1,0]$, i.e.\ using only representations from the first LSTM layer.


\paragraph{Evaluation Scenarios for Dependency Parsing}

For a fair comparison, we use the same setting as used by previous models for each scenario.
Our main model (which we refer to as \textsc{Ours}) is using a \textsc{Supervised Anchored Alignment} (\secref{sec:supervised}) to align the multilingual pretrained ELMo embeddings which are used by the parser.
We compare against several variants of our model:

\begin{itemize}[leftmargin=*]
 
    \item \textsc{Aligned fastText}: instead of ELMo, we use \textsc{fastText} pretrained embeddings~\citep{grave2018learning}, aligned to English using MUSE.
\item \textsc{Aligned $\bar{\ee}$}: instead of contextualized embeddings, we use the anchors themselves as fixed embeddings, aligned to English.
\item \textsc{No Dictionary}: we assume the absence of a dictionary and use \textsc{Unsupervised Anchored Alignment}.
\item \textsc{No POS}: no use of part of speech tags.
\end{itemize}

\section{Results}\label{sec:results}

\paragraph{Alignment} 
As mentioned above, we use outputs of the first LSTM layer of ELMo in our parsing experiments. Therefore, we present the alignment accuracy for those in \tabref{tab:align}, summarizing the precision@5 word-translation from 6 languages to English. Results for the other layers are presented in \appref{appendix:align}. As expected, supervised alignments outperform unsupervised ones by a large margin. Between the two unsupervised methods, the context-based alignment achieved significantly better results for Spanish and Portuguese but failed to converge for Swedish. In both cases, the value of anchors in the \textsc{refine} step is clear, substantially improving the precision for all languages.

\begin{table*}[t!]
\centering
\begin{tabular}{l|l|cc|cc|c}
\toprule
  \multirow{2}{*}{\textsc{\# Sentences}} & \multirow{2}{*}{\textsc{Language Model}} &
      \multicolumn{2}{c}{\textsc{uas / las}} &
      \multicolumn{2}{c}{\textsc{Perplexity}} & \multirow{2}{*}{\textsc{Align}}\\
      & &  \textsc{Dev} & \textsc{Test} & \textsc{Train} & \textsc{Dev} & \\
\midrule
28M          & \textsc{ELMo}                         & 72.3 / 62.8 & 72.5 / 61.3 & 22         & 44       & 85    \\ 
\midrule
 10K         & \textsc{ELMo}                      & 52.9 / 38.3 & 50.1 / 33.1 & 4          & 4060     & 4     \\ 
          & \textsc{Anchored ELMo}         & \textbf{59.2 / 47.3} & \textbf{57.2 / 42.2} & 92         & 600      & 12    \\ 
\bottomrule
\end{tabular}
\caption{Zero-shot, single-source results for the Spanish limited unlabeled data experiments. The parsing results are UAS/LAS scores, the perplexity is of the ELMo model, and the alignment scores are precision@5 on the held-out set, based on CSLS. All embeddings were aligned to English using supervised anchored alignment.}
\label{tab:es_low}
\end{table*}

\begin{table}[t!]
\centering
\begin{tabular}{lc}
\toprule
\textsc{Model}                         & \textsc{LAS-F1}   \\ 
\midrule
\comment{\textsc{CUNI x-ling}} \citet{rosa_cuni_2018}            & 26.31 \\ 
\comment{\textsc{Uppsala}} \citet{smith201882}                & 31.93 \\ 
\textsc{Aligned fastText}                 & 26.77 \\
\textsc{Ours}                 & \textbf{36.98} \\ 
\bottomrule
\end{tabular}
\caption{Results for the Kazakh dataset from CoNLL 2018 Shared Task on Multilingual Parsing, compared to the two leading models w.r.t.\ this treebank.}
\label{tab:kazakh}
\end{table}

\paragraph{Zero-Shot Parsing, Multiple Source Languages} 

Table~\ref{table:zeroshot} summarizes the results for our zero-shot, multi-source experiments on six languages from Google universal dependency treebank version 2.0.\footnote{\url{https://github.com/ryanmcd/uni-dep-tb/}}  For each tested language, the parser was trained on all treebanks in the five other languages and English. We align each of the six languages to English. We compare our model to the performance of previous methods in the same setting (referred to as $L^t\cap L^s=\emptyset$ in \citet{ammar_one_2016}).
The results show that our multilingual parser outperforms all previous parsers with a large margin of 6.8 LAS points. Even with an unsupervised alignment, our model consistently improves over previous models.

To make a fair comparison to previous models, we also use gold POS tags as inputs to our parser. However, for low-resource languages, we might not have access to such labels. Even without the use of POS tags at all, in five out of six languages the score is still higher than previous methods that do consider such annotations. An exception is the Portuguese language where it leads to a drop of 8.8 LAS points. While in the single language setting this good performance can be explained by the knowledge captured in the character level, contextual embeddings ~\citep{smith2018investigation,belinkov2017evaluating}, the results suggest that this knowledge transfers across languages.

In order to assess the value of contextual embeddings, we also evaluate our model using non-contextual embeddings produced by \textsc{fastText}~\cite{bojanowski_enriching_2017}. While these improve over previous works, our context-aware model outperforms them for all six languages in UAS score and for 5 out of 6 languages in LAS score, obtaining an average higher by 3 points. To further examine the impact of introducing context, we run our model with precomputed anchors ($\bar{\ee}$). Unlike \textsc{fastText} embeddings of size 300, these anchors share the same dimension with contextual embeddings but lack the contextual information. Indeed, the context-aware model is consistently better.

\paragraph{Few-Shot Parsing, Small Treebanks} 
In this scenario, we assume a very small tree-bank for the tested language and no POS tags. We use the Kazakh tree-bank from CoNLL 2018 shared task~\cite{zeman-EtAl:2018:K18-2}. The training set consists of only 38 trees and no development set is provided. Segmentation and tokenization are applied using UDPipe. Similar to ~\citet{rosa_cuni_2018, smith201882}, we utilize the available training data in Turkish as it is a related language. To align contextual embeddings, we use a dictionary generated and provided by \citet{rosa_cuni_2018} and compute an alignment from Kazakh to Turkish.
The dictionary was obtained using FastAlign~\cite{fastalign} on the OpenSubtitles2018~\cite{opensub} parallel sentences dataset from OPUS~\cite{opus}.\footnote{\url{https://github.com/CoNLL-UD-2018/CUNI-x-ling}}

\tabref{tab:kazakh} summarizes the results, showing that our algorithm outperforms the best model from the shared task by $5.05$ LAS points and improves by over 10 points over a \textsc{fastText} baseline.

\paragraph{Zero-Shot Parsing, Limited Unlabeled Data}
To evaluate our anchored language model (\secref{sec:anchored_lm}), we simulate a low resource scenario by extracting only 10k random sentences out of the Spanish unlabeled data. We also extract 50k sentences for LM evaluation but perform all computations, such as anchor extraction, on the 10k training data. For a dictionary, we used the 5k training table from \citet{conneau2017word}.\footnote{We filtered words with multiple translations to the most common one by Google Translate.} Another table of size 1,500 was used to evaluate the alignment. In this scenario, we assume a single training language (English) and no usage of POS tags nor any labeled data for the tested language.

Table \ref{tab:es_low} shows the results. Reducing the amount of unlabeled data drastically decreases the precision by around 20 points. The regularization introduced in our anchored LM significantly improves the validation perplexity, leading to a gain of 7 UAS points and 9 LAS points.

\section{Conclusion}
We introduce a novel method for multilingual transfer that utilizes deep contextual embeddings of different languages, pretrained in an unsupervised fashion.
At the core of our methods, we suggest to use anchors for tokens, reducing this problem to context-independent alignment. Our methods are compatible both for cases where a dictionary is present and absent, as well as for low-resource languages. The acquired alignment can be used to improve cross-lingual transfer learning, gaining from the contextual nature of the embeddings. We show that these methods lead to good word translation results, and improve significantly upon state-of-the-art zero-shot and few-shot cross-lingual dependency parsing models. 

In addition, our analysis reveals interesting properties of the context-aware embeddings generated by the ELMo model. Those findings are another step towards understanding the nature of contextual word embeddings. 

As our method is in its core task-independent, we conjecture that it can generalize to other tasks as well.

\section*{Acknowledgements}
We thank the MIT NLP group and the reviewers for their helpful discussion and comments.

The first and third authors were supported by the Office of the Director of National Intelligence (ODNI), Intelligence Advanced Research Projects Activity (IARPA), via contract \# FA8650-17-C-9116. The views and conclusions contained herein are those of the authors and should not be interpreted as necessarily representing the official policies, either expressed or implied, of ODNI, IARPA, or the U.S. Government. The U.S. Government is authorized to reproduce and distribute reprints for governmental purposes notwithstanding any copyright annotation therein.

This work was also supported in part by the US-Israel Binational Science Foundation (BSF, Grant No. 2012330), and by the Yandex Initiative in Machine Learning. 

\bibliography{naaclhlt2019}
\bibliographystyle{acl_natbib}

\appendix
\section{Alignment Results for All Layers}\label{appendix:align}

\tabref{tab:align_all_layers} manifests word-to-word translation results when supervised alignment is performed over different layer outputs from ELMo's LSTM. Even though layer zero produces context independent representations, the anchors computed over the contextual representations achieved higher precision. We conjecture that this is due to the language model objective being applied to the output of the second layer. Hence, unlike token-based embeddings such as \textsc{fastText} that optimize them directly, the context-independent representations of ELMo are optimized to produce a good base for the contextual embeddings that are computed on top of them.

\begin{table*}[t!]
\centering
\begin{tabular}{c|cccccc|c}
\toprule
\textsc{Layer}       & \textsc{de}    & \textsc{es}    & \textsc{fr}    & \textsc{it}    & \textsc{pt}    & \textsc{sv}    & \textsc{average} \\ 
\midrule
0                    & \ \ 54 / 71 \ \  & \ \ 65 / 80\ \  & \ \ 66 / 80 \ \  & \ \ 61 / 78 \ \  & \ \ 55 / 71\ \  & \ \ 41 / 61\ \  &  \ \ 57 / 74\ \    \\ 
1                    & \ \ \textbf{62} / \textbf{78}\ \  & \ \ \textbf{73} / \textbf{85}\ \  & \ \ \textbf{74} / \textbf{86}\ \  & \ \ \textbf{69} / \textbf{82}\ \  & \ \ \textbf{61} / \textbf{74}\ \  & \ \ \textbf{49} / \textbf{68}\ \  &  \ \ \textbf{65} / \textbf{79}\ \    \\ 
2                    & \ \ 59 / 75\ \  & \ \ 68 / 82\ \  & \ \ 70 / 83\ \  & \ \ 66 / 79\ \  & \ \ 56 / 72\ \  & \ \ 48 / 67\ \  &  \ \ 61 / 76\ \    \\ 
\bottomrule
\end{tabular}\\
\caption{Per ELMo layer word translation to English precision @1 / @5 using CSLS~\cite{conneau2017word} with a dictionary (supervised) for German (\textsc{de}), Spanish (\textsc{es}), French (\textsc{fr}), Italian (\textsc{it}), Portuguese (\textsc{pt}) and Swedish (\textsc{sv}). Layer 0 representations are the result of the character-level word embeddings (which are context independent). Layer 1 and 2 alignments are based on anchors from the first and second LSTM layer output respectively.}\label{tab:align_all_layers}
\end{table*}
\section{Additional Parsing Results}\label{appendix:parsing}
In Table \ref{table:appendix_zeroshot} we provide complementary results to those in zero-shot closs-lingual parsing.

\begin{table*}[h!]\label{table:additional_zero}
\begin{tabular}{l|cccccc|c}
\toprule
\textsc{Model}              & \textsc{de}        & \textsc{es}        & \textsc{fr}        & \textsc{it}        & \textsc{pt}        & \textsc{sv}        & \textsc{average}       \\ 
\midrule
\citet{zhang_hierarchical_2015}           & 62.5     & 78.0     & 78.9     & 79.3     & 78.6     & 75.0     & 75.4     \\ 
\citet{guo_representation_2016}                & 65.0     & 79.0     & 77.7     & 78.5     & 81.9     & 78.3     & 76.7     \\ 
\midrule 
\textsc{Aligned fastText} & \ \ 69.2\ \  & \ \ 83.4\ \   & \ \ 84.6\ \  & \ \ 84.3\ \  & \ \ 86.0\ \  & \ \ 80.6\ \  & \ \ 81.4\ \  \\ 
\textsc{Aligned} $\bar{\ee}$       & 65.1 & 82.8 & 83.9 & 83.6 & 83.4 & 82.0 & 80.1 \\ 
\textsc{Ours}               & \textbf{73.7} & \textbf{85.5} & \textbf{87.8} & \textbf{87.0} & \textbf{86.6} & \textbf{84.6} & \textbf{84.2} \\ 
\midrule \midrule
\textsc{Ours, no dictionary}    & 73.2 & 84.3 & 87.0 & 86.8 & 84.5 & 80.4 & 82.7 \\ 
\textsc{Ours, no pos} &   69.7 & 84.8  & 85.3 & 85.3 & 79.7 &  81.7 & 81.1 \\ 
\textsc{Ours, no dictionary, no pos} & 72.2  &	84.7  &	84.9  &	85.0  &	78.1  &	67.9 & 78.8\\ 
\bottomrule
\end{tabular}
\caption{Zero-shot cross lingual results compared to previous methods, measured in UAS. Aligned fastText and $\bar{\ee}$ context-independent models are also presented as baselines. The bottom three rows are models that don't use POS tags at all and/or use an unsupervised anchored alignment. \\
Note that \citet{ammar_one_2016} did not publish UAS results.}\label{table:appendix_zeroshot}
\end{table*}

\section{Hyperparameters}\label{appendix:hyperparameters}

We now detail the hyperparameters used throughout our experiments. All alignment experiments were performed using the default hyperparameters of the MUSE framework (see their github repository). Table \ref{table:hyper} depicts the values used in multilingual parsing experiments. 

\begin{table}[h!]
\centering
\begin{tabular}{lc}
\toprule
\textsc{Hyperparameter} & \textsc{Value} \\
\midrule
\textsc{batch size (\# sentences)} & 32 \\
\textsc{Instances per epoch} & 32,000 \\
\textsc{Epochs (Max.)} & 40 \\
\textsc{Patience (early stopping)} & 10 \\
\textsc{Encoder type} & \textsc{Bi-LSTM} \\
\textsc{POS tag embedding dim.} & 50 \\
\textsc{LSTM, hidden size} & 200 \\
\textsc{LSTM, \# layers}  & 3 \\
\textsc{Dropout rate} & 0.33 \\
\textsc{Arc representation dim.} & 500 \\
\textsc{Tag representation dim.} & 100 \\
\textsc{Adam} & \textsc{(default)} \\

\bottomrule
\end{tabular} 
\qquad
\caption{Hyper-parameters used in parsing experiments, shared across different settings.}
\label{table:hyper}
\end{table}
\section{Additional Alignment Example}\label{appendix:space}

\begin{figure}[t!]
    \centering
  \includegraphics[width=0.51\textwidth]{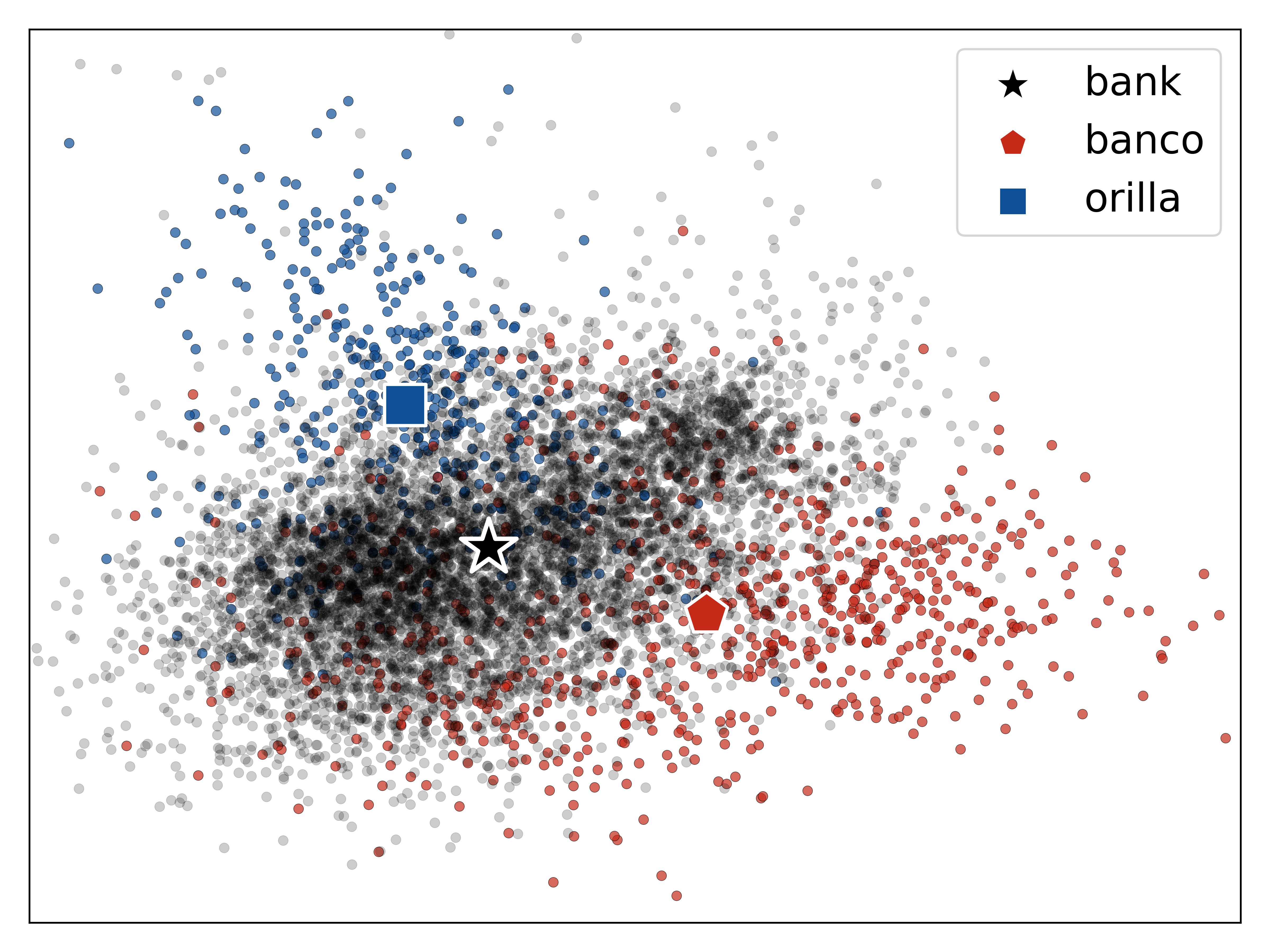}

    \caption{Contextual embeddings for the English word ``bank'' and its two possible translations in Spanish ---  ``banco'' (a financial establishment) in red and  ``orilla'' (shore) in blue. The figure shows a two dimensional PCA for the aligned space of the two languages. The symbols are the anchors and the dots are the contextualized embeddings. (best viewed in color) } \label{fig:bank}
\end{figure}

We provide an additional example of a homonym. \figref{fig:bank} shows the contextual embeddings of the word ``bank'' in English and the words ``banco'' (a financial establishment) and ``orilla'' (shore) in Spanish. In this case, unlike the ``bear'' example (\figref{fig:bear}), the embeddings do not form two obvious clusters in the reduced two dimensional space. A possible explanation is that here the two meanings have the same POS tag (Noun). Even so, as shown in \tabref{tab:examples_bank}, the alignment succeeds to place the embeddings of words from each context close to the matching translation. 

The nearest-neighbors for the ``bear'' example are presented in \tabref{tab:examples_bear}.

\newcommand{\specialcell}[2][c]{%
  \begin{tabular}[#1]{@{}l@{}}#2\end{tabular}}

\begin{table*}[t]
\small
\begin{center}
    \small
    \begin{tabular}{ll}
    \toprule
    $k$ & \textsc{``banco'' anchor}        \\ 
    \hline \addlinespace[0.5ex]
    1 & \specialcell{Unlike in primary succession , the species that dominate secondary succession , are usually present from the start of the\\ process , often in the soil seed \textbf{bank} .}                  \\ \addlinespace[0.5ex]
    2 & \specialcell{Canto XLV is a litany against Usura or usury , which Pound later defined as a charge on credit regardless of potential or\\ actual production and the creation of wealth ex nihilo by a \textbf{bank} to the benefit of its shareholders .}               \\ \addlinespace[0.5ex]
    3 & \specialcell{This prompted some investigation , led by Sir Benjamin Hall , which quickly turned up the fact that O' Connor was\\ registered as the owner of all the estates , and of the associated \textbf{bank} . }           \\ \addlinespace[0.5ex]
    4 & The commercial NaS battery \textbf{bank} offers : ( Japanese ) .                \\ \addlinespace[0.5ex]
    5 & \specialcell{Both team leaders are given a mystery word , which along with their team - mates use gigantic foam blocks and place\\ them on the clue \textbf{bank} ( similar to Boggle ) with only giving a clue to the word ...}               \\ \addlinespace[0.5ex]
    \midrule
    $k$ & \textsc{``orilla'' anchor}        \\ 
    \hline \addlinespace[0.5ex]
    1 & \specialcell{The combined Protestant forces , now numbering 25,000 strong , positioned themselves on the western \textbf{bank} of the Rhine\\ River . }                 \\ \addlinespace[0.5ex]
    2 & The Romans had a small advance guard of auxiliaries and cavalry on the opposite \textbf{bank} of the river .               \\ \addlinespace[0.5ex]
    3 & \specialcell{Between Vaugirard and the river Seine he had a considerable force of cavalry , the front of which was flanked by a battery \\ advantageously posted near Auteuil on the right \textbf{bank} of the river .}            \\ \addlinespace[0.5ex]
    4 & Mallus therefore stood on the eastern \textbf{bank} of the river .                \\ \addlinespace[0.5ex]
    5 & The Argentine squadron spent the night of February 7 anchored between Juncal Island and the west \textbf{bank} of the river . \\ \addlinespace[0.5ex]
    6 & \specialcell{After marching north from Tewkesbury , Sir William Waller tried to contain the cavalry forces of Maurice on the western \\ \textbf{bank} of the Severn , cutting this substantial force off from the rest of the Royalist army .} \\ \addlinespace[0.5ex]

    \bottomrule
    \end{tabular}
    \smallskip
    \caption{Nearest-neighbors (after alignment) of the Spanish anchors ``banco'' (a financial establishment) and ``orilla'' (shore) from the contextual embeddings of the word ``bank'' in English. The full sentence is presented for context.}
    \label{tab:examples_bank}
    \end{center}
\end{table*}

\begin{table*}[t]
\small
\begin{center}
    \small
    \begin{tabular}{ll}
    \toprule
    $k$ & \textsc{``tener'' anchor}        \\
    \hline \addlinespace[0.5ex]
    1 & \specialcell{It may be difficult for the patient to \textbf{bear} the odour of the smoke at first , but once he gets used to such a smell , \\it does not really matter .}                  \\ \addlinespace[0.5ex]
    2 & No matter what the better class of slave owners might do , they had to \textbf{bear} the stigma of cruelty with the worst of tyrants ...               \\ \addlinespace[0.5ex]
    3 & Every new car will \textbf{bear} the Élan name instead of Van Diemen , so the highly successful marque will gradually disappear .            \\ \addlinespace[0.5ex]
    4 & had a sufficient economic stake to \textbf{bear} the litigation burden necessary to maintain a private suit for  recovery under section 4 .                \\ \addlinespace[0.5ex]
    5 & In this example , consumers \textbf{bear} the entire burden of the tax ; the tax incidence falls on consumers . \\ \addlinespace[0.5ex]
    
        \midrule
        
    $k$ & \textsc{``oso'' anchor}        \\
    \hline \addlinespace[0.5ex]
    1 & In 2010 , the government of the NWT decided to update its version of the polar \textbf{bear} - shaped plate .                  \\ \addlinespace[0.5ex]
    2 & \specialcell{Salad Fingers appears to be masochistic , as he can be seen taking pleasure from impaling his finger on a nail , \\ rubbing stinging nettles on himself or stepping onto a \textbf{bear} trap .}               \\ \addlinespace[0.5ex]
    3 & The old \textbf{bear} - hunter , on being toasted , made a speech to the Texians , replete with his usual dry humor .            \\ \addlinespace[0.5ex]
    4 & Balto arrives , distracts the \textbf{bear} , saves Aleu , they both escape and the \textbf{bear} disappears .                \\ \addlinespace[0.5ex]
    5 & Defeated , the polar \textbf{bear} shrinks and transforms into a plush toy .               \\ \addlinespace[0.5ex]

    \bottomrule
    \end{tabular}
    \smallskip
    \caption{Nearest-neighbors (after alignment) of the Spanish anchors ``tener'' (carry, verb) and ``oso'' (animal, noun) from the contextual embeddings of the word ``bear'' in English.  The full sentence is presented for context.}
    \label{tab:examples_bear}
    \end{center}
\end{table*}

\end{document}